\documentclass[sigconf]{acmart}
\AtBeginDocument{%
  \providecommand\BibTeX{{%
    \normalfont B\kern-0.5em{\scshape i\kern-0.25em b}\kern-0.8em\TeX}}}
\acmYear{2020}  
\copyrightyear{2020}  
\setcopyright{acmcopyright} 
\acmConference{AIIS'20}{July 30, 2020}{Virtual Event, China}  
\acmPrice{TBA}  
\acmDOI{TBA}  
\acmISBN{TBA}
\begin{document}

%%
%% The "title" command has an optional parameter,
%% allowing the author to define a "short title" to be used in page headers.
\title{From A Glance to ``Gotcha'': Interactive Facial Image Retrieval with Progressive Relevance Feedback}

%%
%% The "author" command and its associated commands are used to define
%% the authors and their affiliations.
%% Of note is the shared affiliation of the first two authors, and the
%% "authornote" and "authornotemark" commands
%% used to denote shared contribution to the research.

%%
%% By default, the full list of authors will be used in the page
%% headers. Often, this list is too long, and will overlap
%% other information printed in the page headers. This command allows
%% the author to define a more concise list
%% of authors' names for this purpose.

\author{Xinru Yang}
\affiliation{\institution{Carnegie Mellon University, Google LLC}}
\email{xinruy@cs.cmu.edu}

\author{Haozhi Qi}
\affiliation{\institution{UC Berkeley}}
\email{hqi@berkeley.edu}

\author{Mingyang Li}
\affiliation{\institution{University of Pennsylvania}}
\email{myli@alumni.upenn.edu}

\author{Alexander Hauptmann}
\affiliation{\institution{Carnegie Mellon University}}
\email{alex@cs.cmu.edu}

%%
%% The abstract is a short summary of the work to be presented in the
%% article.
\begin{abstract}
  Facial image retrieval plays a significant role in forensic investigations where an untrained witness tries to identify a suspect from a massive pool of images. However, due to the difficulties in describing human facial appearances verbally and directly, people naturally tend to depict by referring to well-known existing images and comparing specific areas of faces with them and it is also challenging to provide complete comparison at each time. Therefore, we propose an end-to-end framework to retrieve facial images with relevance feedback progressively provided by the witness, enabling an exploitation of history information during multiple rounds and an interactive and iterative approach to retrieving the mental image. With no need of any extra annotations, our model can be applied at the cost of a little response effort. We experiment on \texttt{CelebA}~\cite{liu2015faceattributes} and evaluate the performance by ranking percentile and achieve 99\% under the best setting. Since this topic remains little explored to the best of our knowledge, we hope our work can serve as a stepping stone for further research.
\end{abstract}

%%
%% The code below is generated by the tool at http://dl.acm.org/ccs.cfm.
%% Please copy and paste the code instead of the example below.
%%

%%
%% Keywords. The author(s) should pick words that accurately describe
%% the work being presented. Separate the keywords with commas.

%% A "teaser" image appears between the author and affiliation
%% information and the body of the document, and typically spans the
%% page.
\begin{teaserfigure}
  \includegraphics[width=\textwidth]{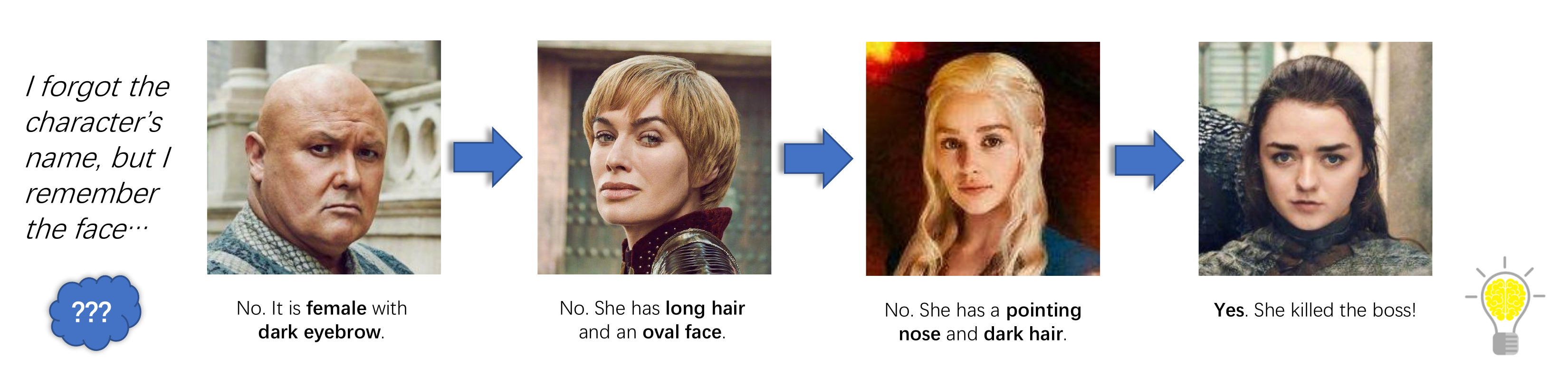}
  \caption{When a user is trying to identify a character by providing relative descriptions for multiple rounds. Images are property of 2019 Home Box Office, Inc., reproduced under fair use.}
  \label{fig:teaser1}
\end{teaserfigure}

%%
%% This command processes the author and affiliation and title
%% information and builds the first part of the formatted document.
\maketitle
%%%%%%%%% BODY TEXT
\section{Introduction}
%scenario & assumptions & problem conceptualization
Facial image retrieval is an interesting yet challenging task for its practical use in computational forensics. It is even more difficult when the target image is not known to the system but only exists in the user's mind (mental image). Therefore, the descriptions of the target image from the user are necessary for a reasonable retrieval results. However, compared to describing the perceived face with absolute depictions, people naturally feel it easier to refer to an existing image and provide the descriptions of the their difference. For example, in Figure~\ref{fig:teaser1}, when a user is trying to identify a character without knowing the character's name but only an impression of the character's face, a reference image is shown to the user and the user responds with differences between the reference image and the mental image. Upon receiving these responses, another reference image will be shown to the user for further feedback. Ideally, the reference images shown to the user should be refined and closer to the mental image over time. This process might last for limited rounds or till the mental image is retrieved. 
%\hqi{I think I have a misunderstanding here... What is the meaning of image is not known to the system? My understanding is the image is not in the database. But it seems to be wrong since the image cannot be retrieved if it's not known..}. %narration from the user's view.
%reference - candidate; mental - target;
%the system first presents a candidate image and gathers relative descriptions from the user. Then it refines the results based on the given candidate image and the feedback and retrieves another candidate image and presents it to the user again. This process might last for several rounds until the system ultimately retrieves the correct one.

%\begin{figure*}[htb]
%    \centering
%    \includegraphics[scale=0.5]{teaser0.pdf}
%    \caption{When a user is trying to identify a character by providing relative descriptions for multiple rounds.}
%    \label{fig:teaser0}
%\end{figure*}

%why use attributes
In reality, when people are asked to describe facial appearances of another person, the descriptions can be roughly categorized into two aspects, basic descriptions and advanced descriptions. %\hqi{How about directly calling objective/subjective here?}
Basic descriptions contain objective measurements on facts such as the color of hair, whether wearing hat or eyeglasses, which can be conveniently mapped into attributes with certain values. Advanced descriptions involve with subjective opinions people sense from the facial appearances such as beauty, aging, friendliness etc. Previous studies have demonstrated the instability in advanced descriptions due to the ambiguities in human perception~\cite{147470491301100414,cshwlaf,mfriuac,engelmann2013emotion}. %Moreover, the lack of universality makes numerical descriptors, both absolute (``eye width 2.6 cm'') and relative (``nose should be 2 times bigger''), unsuitable for facial image retrieval.
Moreover, researchers have expressed concerns that verbally attending to facial differences might alter witness's memory of the original face, which can be detrimental to forensic applications~\cite{brown2005verbal}. Therefore, we take an attribute-aware approach (e.g. ``with or without glasses'') where users are able to describe and response easily and efficiently. An appropriate dataset for this research goal is \texttt{CelebA}, which will be described in Subsection \ref{subsec:dataset}.

%Studying facial features presents extra challenges due to ambiguities in human perception. For example, different cultures may define beauty differently~\cite{147470491301100414}, prioritize facial features differently~\cite{cshwlaf,mfriuac}, and interpret different emotions from same facial expressions~\cite{engelmann2013emotion}.

%\paragraph{Facial Features Are Difficult to Verbally Describe.}
%Compared to manufactured products, faces are more ambiguous when described verbally. Prior studies exploring facial image generation with verbal descriptions did not consider subjective opinions (old, happy, pale, etc.) or relative descriptors~\cite{benhidour2008interactive}. 

%With these challenges considered, we choose to implement a visual RA system, instead of relying on free-form, natural language feedback as demonstrated in prior work~\cite{guo2018dialog}.

%our model applies an attribute-aware method that by comparing specific areas of faces, user is able to provide the relevance feedback easily and efficiently.

%advantages: interaction. compared to previous work
Recent research ~\cite{Kovashka2015,4409072,Zavesky:2008:CEF:1460096.1460136} shows that interactive image retrieval has the advantages of integrating user feedback and improving retrieval performances by relevance feedback. Therefore, we build an interactive retrieval system that takes multiple rounds of collecting user's feedback and refining the retrieval results.
Apart from the interactive framework, our model considers an extra mechanism, progressiveness, during the interactions.

%advantages: progressive. 
Considering the difficulty for real people in describing and comparing facial images in mind, it is hard for them to provide thorough feedback at each round. Thus, we design a mechanism that progressively discloses the feedback. Specifically, in each round of retrieval, the system only provides partial relevance by masking the rest of it. In the following rounds, the ratio of masked relevance feedback is gradually decreased, allowing for more information to be disclosed. This setting mimics a progressive disclosure that might better reflect the functionality of human memory.
Essentially, it outperforms all other settings in our experiments and might share a similar fundamentality as dropout.

%little efforts of extra annotations
With the target image unknown to the retrieval system but only its attributes, one naive way would be annotating the attributes of every image in the database and seeking for the closest one. However, annotating a dataset is highly expensive and exhaustive. Thus, our model retrieves the image by cooperating image features and instant feedback without prior annotations. When the system returns a candidate image in each round of retrieval, the user is only required to provide relevance feedback between it and the mental target image. We believe this instantaneous responding process is light and feasible since our system has only a few rounds with only $1$ image each round.

% Why this is important? 
%A shoplifter fled away, leaving no witness but you. 
%Closed-circuit television (CCTV) was enabled in the perimeter but not inside the store, so you had to help the securities identify the thief from $10$ hours' worth of video footage.
% Why this is hard?
%Another day in life was going to be wasted, you thought.
% 1) Witnesses are seldom trained for accurately describing a face seen.
%Neither the securities nor you, an average customer, were trained for facial compositing or any other forensic techniques. 
% 2) Time consuming to browse through thousands hours of CCTV recordings.
%Also, the securities were paid for hours watching the recordings, but not you.

% How to solve the problem?
%Thankfully, to make your life easier, \textbf{the securities mediated between you and the CCTV}: 
%They pointed at some face on the screen, and they asked you if it looked like the thief.
%You responded ``\textit{the nose is too big}'', ``\textit{yeah the thief was bald too}'', etc., essentially \textbf{describing the faces and providing comparisons}.
%As the conversation went on, the securities learned the way you describe appearances, which made the search more efficient.
%In less than 5 rounds of retrieval, images of the thief was found.

The key contributions of our work are as follows: %= =|| tried my best
\begin{itemize}
    \item A new retrieval problem setting on human facial images under interactive search where users are allowed to convey their mental images to the system and iteratively refine the retrieved results.
    \item An end-to-end interactive Content-Based Image Retrieval (CBIR) framework to address the above problem setting by employing supervised learning approach.
    \item A novel progressive disclosure mechanism in collecting relevance feedback from users during multiple rounds of interaction. The mechanism reaches the best performance while mimicking human behaviors.
    \item An instant feedback setting for interactive applications. The setting can help to reduce workload of manual annotations necessary for learning about the annotator.
    
    % \item We conduct multiple experiments on the \texttt{CelebA} dataset to validate the advantage of our model. 
\end{itemize}

%\subsection{Organization}
The paper is organized in $5$ sections.
After reviewing related work in Section \ref{sec:related}, %we first introduce the key related work to this research, and then we introduce how human describe facial images during their interaction with the environment.  Based on this, 
we describe the structure of our framework (in Subsection \ref{subsec:structure}) and the algorithm it runs by (in Subsection \ref{subsec:algorithm}) in Section \ref{sec:method}. 
%Then we show the settings and results of our experiments followed by analysis and interpretation.
We then continue to Section \ref{sec:experiment}, where we demonstrate the validity with a baseline experiment and the robustness with a series of ablation studies.
%demonstrate the validity of our design with a benchmark experiment (Subsection \ref{subsec:dataset}), the \textbf{robustness} with a \textit{progressive disclosure} experiment (Subsection \ref{subsec:progressive}), and the \textbf{efficiency} with a \textit{baseline} experiment (Subsection \ref{subsec:baseline}).
% Finally, we conclude the work and discuss about the future research prospects for human facial images retrieval.
Finally in Section \ref{sec:conclusion}, we make final remarks about our work in terms of its applications and limitations.

\section{Related Work}\label{sec:related}
In this section, we introduce related work in the fields of image retrieval and facial recognition. 

\subsection{Image Retrieval Researches}

\paragraph{Overview.}
With countless images generated everyday, efficient navigation demands intuitive approaches that are aware to image content, giving rise to the field of Content-Based Image Retrieval (CBIR)~\cite{410145}.
To further facilitate retrieval, interactive querying methods are being developed over the past decades. A prominent approach in this area is relevance feedback (RF)~\cite{Rui1998RelevanceFA}. In traditional RF settings, users evaluate how relevant one retrieved image is to their desired result, report this perceived relevance as a numerical value, and expect a refined result from the next retrieval.

\paragraph{Relative Attributes.}
However, for complex images, a single relevance value can be ambiguous and thus misleading~\cite{geman2000stochastic}. In order to improve specificity in feedback, relative attributes (RA) has been proposed as a new mechanism~\cite{kovashka2012whittlesearch}. In RA-enabled CBIR, user dictates which attribute(s) of candidate image(s) should be tweaked, and optionally by how much. Users may also tune the parameters of the system with an emphasis on specific attributes and image features~\cite{410146,Ma1999,1234821}.

\paragraph{Interactive Feedback.}
Feedback mechanism implies that each retrieval involves multiple rounds of information exchange, or a \textit{dialog}.
Each round provides the CBIR system with extra information to refine its results.
This refinement process can take the form of a decision tree~\cite{macarthur2002interactive} or a neural network (NN)~\cite{wang2006relevance}. In NN-based implementations, reinforcement learning (RL) %has been proved capable of training networks with reduced human supervision
can be employed to reduce training-time supervision~\cite{das2017learning}.
Owing to the descriptive nature of feedback, CBIR experience can be further enhanced with natural language processing (NLP)~\cite{1997OptEn}.
Moreover, a combination of RL and NLP is proved useful in a setting of shoe-shopping~\cite{guo2018dialog}. %\hqi{should we discussed the difference with them since they are the most related one?}%Our model is partially implemented based on this prior work.

\subsection{Facial Image Researches}
\label{subsec:facerecog}

\paragraph{Use Case.}
In our work, we study the retrieval of human face images instead of footwear. 
Face image retrieval has socially critical applications in industries such as forensics~\cite{monroe2009method}.
A typical use case would be having a witness identify the appearance of some specific personnel from Closed-circuit television (CCTV) recordings. The footage repository can be huge, impossible for untrained eyes to examine thoroughly. This workload demands facial recognition technology combined with CBIR techniques.

\paragraph{Feature Extractor.}
%TODO: similar to NIPS
Plenty of CBIR research involving RF has focused on algorithmically-generated visual descriptors, such as \texttt{MPEG-7}~\cite{wong2005mirror}. These low-level features (hue, angle, slope, etc.) are infamously difficult to map to high-level concepts (glasses on, oval face, etc.)~\cite{rui1997relevance}. To bridge over this gap, our pipeline employs a pre-trained Convolutional Neural Network (CNN) for extracting facial features. 

\section{Method}
\label{sec:method}
\subsection{Model Architecture}
\label{subsec:structure}

\begin{figure*}[!tb]
    \centering
    \includegraphics[scale=0.5]{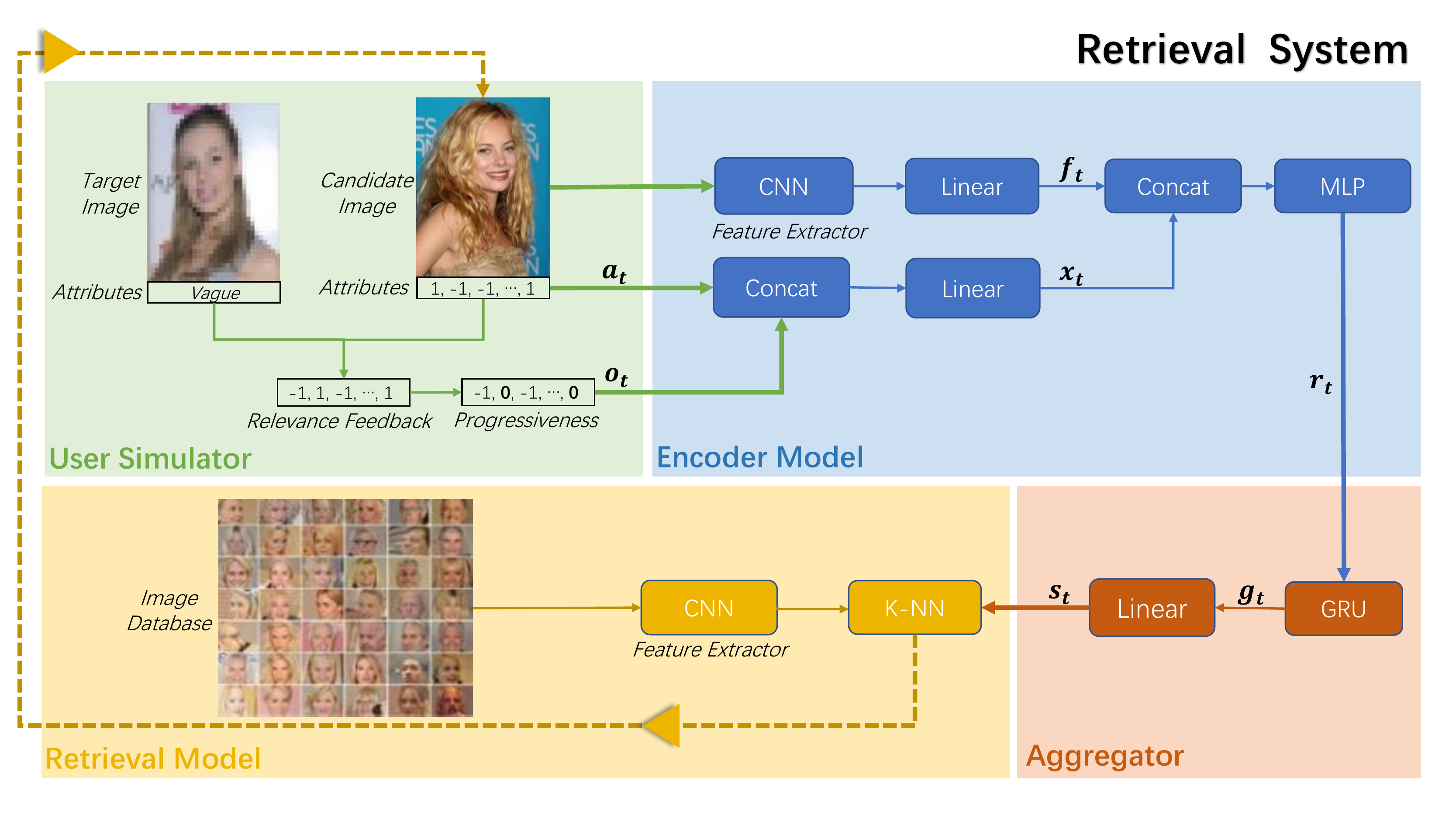}
    \caption{Our proposed end-to-end interactive facial image retrieval framework with progressive disclosure.}
    \label{fig:pipeline}
\end{figure*}

%Our model comprises a database and a retrieval system. All images in the database are pre-processed with a feature extractor to obtain a corresponding feature representation. The system contains four parts, an encoder model and a retrieval model. To automate tests, the system is added with a User Simulator, which will not be present in production. We refer to the combination of the system with the US as \textit{the pipeline}, shown in Figure \ref{fig:pipeline}.

Our model, shown in Figure \ref{fig:pipeline}, consists of four components for different purposes. They will be demonstrated in detail respectively as follows:

\subsubsection{The User Simulator}
% User.
The user simulator mimics a human user who has a target image in mind and provides feedback at each round.

A human user 1) annotates and attributes from a given candidate image, 2) compares specific attributes of the candidate image with those of the target image (which only exists in the user's mind), and 3) reports to the encoder model. Note that the target image is not known to the system because it symbolizes the image in the memory of the witness and its attributes are not explicitly defined either. %The user is also expected to have the target image features annotated beforehand.

%To automate experiments in Section \ref{sec:experiment}, we replace human users with a simulator. The annotation task requires the US to incorporate a facial feature extractor, such as \texttt{SE-ResNet}.
In training process, we utilize the existing annotations of attributes in \texttt{CelebA} to avoid additional inaccuracies if a new annotator is introduced.
In testing, we assume that the online annotations by a person should be the same as the existing annotations under ideal circumstances. Therefore we again utilize the existing annotations to examine our model in this case.

\subsubsection{The Encoder Model}
% What does the EM do? What's its purpose?
This encoder model encodes signals from different spaces into one unified representation.
%The EM serves the RM by embedding all relevant information into a single vector.
Besides the candidate image annotations and the relevance feedback provided by the user simulator, the candidate image itself is also referenced so the encoder model can learn the correspondences between image features and attributes to produce a shared representation of them.
%% What are those signals?
%Such information includes:
%1) the relevance feedback sent from the US, 
%2) the binary attributes of the candidate image,
%3) the candidate image itself, and % (a 224x178x3 tensor) 
%4) the retrival history.% into a shared representation space. This is done as follows:
% How is the *candidate image* utilized?
%To extract low-level (thus real-valued) facial features from the candidate image, a convolutional neural network (CNN) is employed. %For coherency with database pre-processing, we choose \texttt{SE-ResNet} as the CNN and freeze its weights.

\subsubsection{The Aggregator}
The aggregator aggregates the history representation during multiple rounds to exploit the information of previous rounds.
% How is the *history* utilized?
Specifically, a gated recurrent unit (GRU) is implemented followed by a linear transformation.

\subsubsection{The Retrieval Model}
The retrieval model searches the database for a new candidate image that best matches the representation and returns it back to the user.

Upon receiving, it computes the distance between the representations of each image in the database and the aggregated representations from aggregator and selects the nearest neighbors.

During training, it returns a random image in these nearest neighbors for the sake of robustness; In testing, we use greedy approach that returns exactly the nearest one.

\subsection{Algorithm}
\label{subsec:algorithm}
\paragraph{Initialization.}
At the beginning of each retrieval, the user simulator randomly selects an image from the database as its target image. 
The user simulator then annotates and stores the target image for the convenience of calculating relevance feedback in each round. For simplification, we use the existing annotations in the dataset which is a $40$-dimensional Boolean vector $\vec{q} \in \mathbb{R}^{40}$ where $\mathbb{R}=\{-1,1\}$.
Before the first round of retrieval, retrieval model randomly returns a candidate image from the database as a starting point. %Meanwhile, EM is initialized to a random state, denoted by $\vec{s}_0$. 

\paragraph{Loop.}
After initialization, the system executes the following steps iteratively until termination.

\subparagraph{In the User Simulator.}
At the $t$-th round of retrieval, the user simulator first annotates the $t$-th candidate image and stores as a $40$-dimensional Boolean vector $\vec{a}_t \in \mathbb{R}^{40}$. 
Then it calculates the relevance $\vec{o}_t$ between the $t$-th candidate image and the target image: $\vec{o}_t=\vec{a}_t \circ \vec{q} \in \mathbb{R}^{40}$, where $\circ$ denotes elementwise multiplication.
The attributes, together with all other Boolean values, takes $-1$ as \texttt{False} and $1$ as \texttt{True}.
This enables the user simulator to \textbf{calculate the relevance} between the attributes of candidate image and those of target image and the relevance is also binary using the same numbers as attributes:
A term in the relevance is $-1$ if the corresponding attribute in candidate image is different than that in the target image and $1$ if they are identical.
%progressive
To realize the progressiveness during the retrieval, the relevance feedback will be replaced by $0$ in accordance with certain proportion $p_t$ representing the masked part. As the $t$ increases, $p_t$ will gradually decrease indicating more and more disclosure in the relevance feedback. In our implementation, we set $p=\{0.5, 0.3, 0.2, 0.1, 0.0\}$.
The computed relevance $\vec{o}_t$ is then fed to the encoder model along with the annotated attributes $\vec{a}_t$.

\paragraph{\textit{In the Encoder Model.}}
Firstly, $\vec{o}_t$ and $\vec{a}_t$ are concatenated together and embedded by a linear transformation named \textit{indication layer}, with the intuition that some attributes (such as gender, compared to nose size) are more indicative than the others: $\vec{x}_t = W_A(\vec{o}_t \bigoplus \vec{q})$. Here, $\bigoplus$ denotes concatenation, and $W_A \in \mathbb{R}^{80\times256}$ is our first linear transformation.
Meanwhile, a CNN $Conv$ is employed to extract the features of the candidate image which is passed through another linear transformation: $\vec{f_t} = W_I(Conv(Candidate Image))$.
In our implementation, $W_I \in \mathbb{R}^{256\times256}$, and $Conv$ is a pre-trained SE-ResNet~\cite{hu2018senet}.
Outputs from the these two linear transformations % the results of feature extraction and attributes indication 
are concatenated together and fused in a multi-layer perceptron (MLP): $\vec{r_t} = W_M(\vec{x_t} \bigoplus \vec{f_t})$, where $W_M \in \mathbb{R}^{512\times256}$ is the MLP. % Once we acquire the representations $\vec{f_t}$ of candidate image and the relevance of attributes $\vec{x_t}$, we concatenate them together and project them into a shared representation space 

\paragraph{\textit{In the Aggregator.}}
Historical information is then referenced in a GRU followed by a third linear transformation, $W_G \in \mathbb{R}^{256\times256}$: % To further exploit the history information in previous turns, we employ a GRU followed by a third LT: %, combined with historical information using the GRU, passed through a final LT, and sent to the RM. % model to obtain the final representation. Since our system runs multiple turns, we employ a gated recurrent unit (GRU) to aggregate the history of previous turns during the whole process. The aggregated representation will then be passed through a third LT and reach the final component of our system, the RM.
$\vec{g_t}, \vec{h_t} = GRU(\vec{r_t}, \vec{h_{t-1}}), \vec{s_t}=W_G\vec{g_t}$, where hidden state $\vec{h_t} \in \mathbb{R}^{256}$ and the output of GRU $\vec{g_t}\in \mathbb{R}^{256}$. The final representation, $\vec{s_t} \in \mathbb{R}^{256}$, consists of history representations and information of the current round.

\paragraph{\textit{In the Retrieval Model.}}
Next, $\vec{s_t}$ is sent to the retrieval model. 
For $\forall i=1,2,..., N$ where $N$ denotes the size of the database, it calculates the $L2$ distance between $\vec{s}_t$ and $\vec{m}_i = Conv(I_i)$, the feature representation the $i$-th image $I_i$: $d_i=\lVert \vec{s}_t -\vec{m}_i \rVert_2$.
Using the $L2$ distances, the top-$K$ nearest neighbors of $\vec{s_t}$ can be found, denoted by $\vec{n}\in\mathbb{N}^{K}$.
We model the sampling probability with a softmax distribution over the top-$K$ nearest neighbors:
\begin{equation}
    \pi(j)=\mathrm{e}^{-d_j} / \sum^K_{k=1}\mathrm{e}^{-d_{n_k}}, j = 1,2,\dots,K.
\end{equation}

%Base on $K$ nearest images and their sampling probabilities respectively, 
Two approaches can be adopted to choose the $(t+1)$-th candidate image $I_j'$: %retrieve an image, $I_j'$, as the new candidate image that will be sent to US for the next turn:
\begin{itemize}
    \item In training, we choose a random image $I_j'$ where $j'\sim\pi$.
    \item During testing, we choose the nearest image $I_j'$ where $j'=\arg_{j}\max\pi(j)$.
\end{itemize}

\paragraph{Termination.}
The loop terminates when user simulator reports that candidate image \textit{is} target image, or when the maximum number of rounds (default $5$) is reached.

\subsection{End-to-end Training}
In practice, the system might return multiple candidate images for the user in each turn and collect their relevance feedback respectively for better retrieval performance. While in our work, we simplify the scenario by returning a single image in each turn. It is also available to extend our framework to the practical case by enabling the user to choose one preferred image out of multiple candidate images to obtain the relevance feedback.

%It is challenging and expensive to annotate the relevance feedback of images in such a large dataset offline. Thus, in each turn of retrieval, we adopt a user simulator that substitutes the role of a real user to provide the relevance feedback. In our model, the user simulator directly refers to the pre-defined attributes of \texttt{CelebA}.

Aiming at improving the ranking position of the target image, We train the model by a supervised learning objective. In the beginning, all the parameters of the network are randomly initialized. For loss function, we refer to ~\cite{guo2018dialog} where it uses triplet loss objective.
\begin{equation}
    \mathcal{L} = \mathbb{E}[\sum^T_{t=1}max(0, \left \|\vec{s_t}-\vec{x^+} \right \|_2 - \left \|\vec{s_t}-\vec{x^-} \right \|_2 + m)]
\end{equation}
where $\vec{x^+}$ is the features of the target image and $\vec{x^-}$ is the features of a random image sampled from the database as a negative sample. $m$ is a hyper-parameter and constant representing the margin. $\left \|.\right \|$ means $L2-$norm. Even though the ranking position is not available to learn directly since it is not differentiable, we can exploit the advantage of triplet loss objective that the rank of the target image can be improved by ensuring the proximity of the target image and candidate images.

As for evaluation, we report the average ranking percentile of all the image in the training or testing set. More details on ranking percentile will be described later.

\section{Experiments}
\label{sec:experiment}
%TODO
All experiments are conducted on a NVIDIA 1080 Ti GPU and it takes about 14 hours for training 14 epochs. We implement the framework partially based on ~\cite{guo2018dialog}.

\subsection{Dataset}
\label{subsec:dataset}
\begin{figure}
    \centering
    \includegraphics[width=\columnwidth]{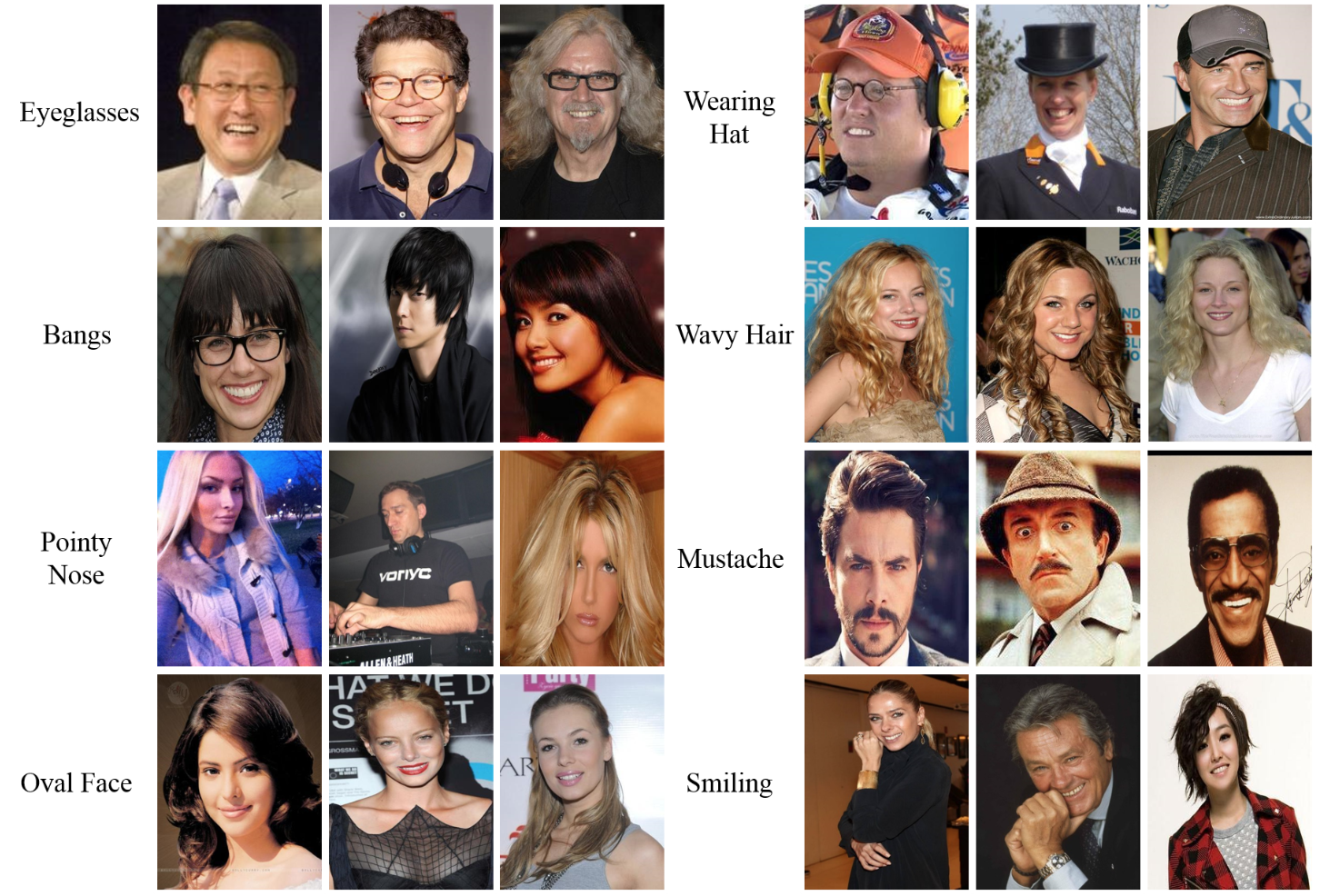}
    \caption{Sample images from the \texttt{CelebA} dataset.}
    \label{fig:dataset}
\end{figure}

\subsubsection{Statistics.}
We employ the \texttt{CelebA} dataset for benchmark purposes. \texttt{CelebA} contains $202,599$ facial images from $10,177$ identities. It is about $13$ times larger than the 
\texttt{Shoes} dataset ($14,765$) studied in~\cite{guo2018dialog}. Each face image is labeled with $40$ binary attributes, such as ``big nose'' and ``bald''.% and marked with $5$ landmark locations.
The dataset contains $115,114$ unique combinations of attributes and the top 10 frequency of identical set of attributes are $358, 203, 200, 184, 182, 181, 147, 143, 142, 137$. However, due to the different poses and angles the same person might have on different images, we use the whole dataset to train and test our model.

\subsubsection{Reasons to Use \texttt{CelebA}}
Using attributes can help avoid ambiguity in human perception as discussed in Subsection \ref{subsec:facerecog}. \texttt{CelebA} also covers various ethnicities and genders, making it popular among facial image researchers~\cite{zhang2016gender,guccluturk2016convolutional}. Its binary attributes, massive coverage, and wide acceptance made the dataset a sufficient choice for our purposes.

\subsection{Model Setup}
We first experiment with different selections of hyper-parameters in Table~\ref{exp:hyper}, and we use the best selection where the constant margin $m$ is set to $2.0$. Then, we experiment with data pre-proscessing and reshape the images from $218\times178$ to $308\times256$ and zero-center them before extracting their features.
We use the first $180,000$ images sorted by name of the files as training set and the rest as testing set. There are $852$ ($8.37\%$ of all $10,177$ identities) individuals whose images will appear in both training set and testing set. The number of rounds is fixed and set to $5$. The learning rate is set to $0.001$. We use adam~\cite{adam} as optimizer and the value of weight decay is set to $0.0$.

\subsection{Metrics}
The metric is the ranking percentile directly referred from~\cite{guo2018dialog}. The reason why we do not use precision is that for each query (target image), there is only one relevant answer in a huge pool (~$200,000$). Unlike QA system or search engines where multiple items can be labeled as relevant, this task is challenging because it asks for highly refined retrieval to get the single relevant answer. Therefore, we calculate the ranking percentile of the target image in the whole search space by their $L2$ distance of representations (which are processed offline using SE-ResNet model). Note that even if there are some images have exactly the same attributes with the target images and we rank them top, we only count the ranking percentile of the target image, which might be lower. The higher percentile is, the more accurate the model is, and the more likely the model can retrieve the correct target image (even though there might be some images with the same attributes that are not target image). We don’t do per-attributes evaluation. And more images share the same attributes will only degrade the model performance rather than cheating or make the number higher. 

\subsection{Baseline}
\label{subsec:baseline}
%TODO???
Attribute-based Retrieval: To retrieve a target image in mind, we can refer to its attributes and search in our database and return an image with closest description. Since in our scenario, the attributes of images are not known in advance, so we split the dataset to train a forty-dimensional classifier on the training set and test its performance on the test set. Note that the training set and testing set are the same for baseline experiment for fair comparison. The evaluation metrics are a bit different that instead of calculating $L2-$ distance, we sort the images in the database based on the number of matched attributes between them and the target image and report the ranking position of it. 

\subsection{Ablation Studies}
\label{subsec:ablation}
\subsubsection{Different Types of Input}
\begin{itemize}
    \item Full disclosure with attributes of candidate image: Unlike progressive disclosure, this setting reflects an extreme situation where the thorough disclosure is provided by a complete comparison between the target image and candidate image at each round.
    %TODO more rounds of experiments.
    \item Full disclosure without attributes of candidate image: Instead of encoding the attributes of candidate image and the relevance feedback together, we experiment with the absence of attributes information of reference image. This setting can reduce the cost of providing feedback for real users since they only need to say ``unlike'' or ``like'' rather than ``unlike the big nose'' or ``like the curly hair''.
    \item Progressive disclosure: As described in our work.
\end{itemize}

\subsubsection{Different Features of Images}
Instead of training and learning the features dynamically, we employ the following pre-trained models to extract the features of images to save us a lot of computational cost.
\begin{itemize}
    \item OpenFace~\cite{amos2016openface}: A face recognition model trained on Labeled Faces in the Wild (\texttt{LFW}) dataset ~\cite{LFWTech}. The features will be extracted to a 128-dimensional representations.
    \item SE-ResNet on \texttt{VGGFace2}~\cite{cao2018vggface2}: A face recognition model trained on \texttt{VGGFace2} dataset. The features will be extracted to a 256-dimensional representations.
    \item SE-ResNet on \texttt{CelebA}: Based on SE-ResNet on \texttt{VGGFace2}, this model is fine-tuned on \texttt{CelebA} to classify $40$ attributes. Apart from designing it as our baseline experiment, we extract the features from its last layer in our model which is a 256-dimensional representations.
\end{itemize}

\subsection{Results and Analysis}
\subsubsection{Baseline and Our Model}
Realistically, there might be multiple samples that share the same combination of attributes, they will be sorted into a consecutive sequence in the database. In this case, retrieving any of them would be considered as a valid operation for the system. Thus, we report in Table \ref{exp:feat} the position of the head and tail of the sequence as the upper bound and lower bound respectively and calculate the expectation by mean value. Note that all other settings are the same and the best.
\begin{table}[h]
    \centering
    \begin{tabular}{c|c|c|c}
        Method & Upper Bound & Lower Bound & Expectation \\
        \hline
        Baseline    & 98.79\%  &   97.09\%  &  98.19\%  \\
        Ours    & N/A & N/A & \textbf{98.66\%}
    \end{tabular}
    \caption{Ranking percentile for baseline and our model.}
    \label{exp:feat}
\end{table}

The results demonstrate a better performance from our model. Though the numbers might look close to each other, the absolute difference, $0.47\%$, will be amplified by the enormous number of image pool. For large dataset such as \texttt{CelebA} which contains more than $200,000$ images, this absolute difference means more than $940$ images are examined and excluded as irrelevant samples in our model. We believe this is of significant importance for users to save their efforts of looking at more than $940$ images unnecessarily.

\subsubsection{Explorations of Hyper-parameters}
We experiment on the value of margin $m$ in the loss objective. Also we experiment with different ways of pre-processing the images to obtain the best performance. Note that the feature extractor we use here is SE-ResNet on \texttt{VGGFace2} and full disclosure is provided in each round. When experimenting with margin, we set reshaped size as $214*178$. When experimenting with reshaped size, we set margin as $2.0$.

\begin{table}[h]
    \centering
    \begin{tabular}{c|l||c|l}
        Margin & Percentile & Reshaped Size & Percentile \\
        \hline
        2.0    & \textbf{92.82\%} &214*178 (raw) & 92.82\%            \\
        1.0    & 92.43\% & 385*320       & 93.21\% \\
        0.5    & 92.21\% & 308*256       & \textbf{93.40\%}   \\
        0.1    & 88.16\%  & 224*224 (crop)  & 91.57\%
    \end{tabular}
    \caption{Ranking percentile is reported on testing set on the final round of the best epoch.}
    \label{exp:hyper}
\end{table}
\subsubsection{Different Choices of Feedback}
\begin{figure}[htb]
    \centering
    \includegraphics[scale=0.67]{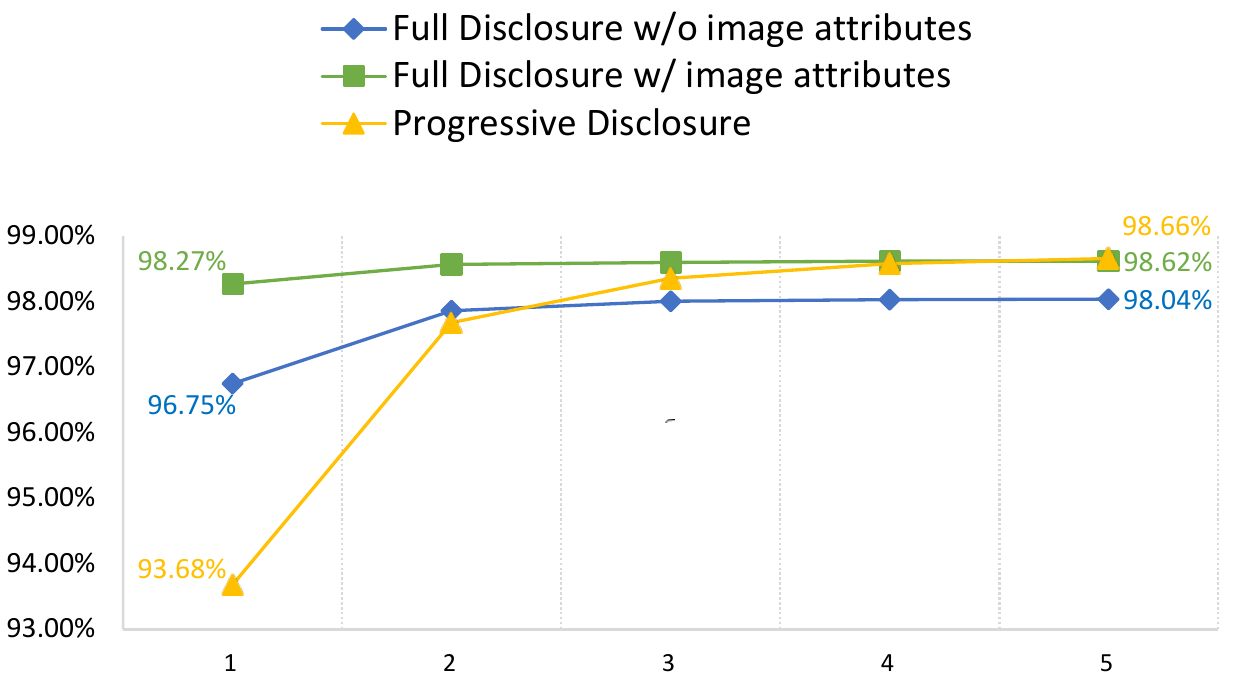}
    \caption{Ranking percentile on testing set in 5 rounds under different settings: full disclosure with and without attributes of candidate image and progressive disclosure.}
    \label{exp:FFF}
\end{figure}

As shown in Figure ~\ref{exp:FFF}, without using attributes of candidate image, the results is limited. While applying progressive disclosure fails to utilize information completely in the beginning, as the disclosure is progressively enhanced, it ultimately outperforms slightly better than full disclosure with attributes of candidate image. When using full disclosure at each round, the growth in ranking percentile soon stagnates at the second round. On the contrary, using progressive disclosure continuously climb and does not converge till the fourth round.
\begin{figure}[htb]
    \centering
    \includegraphics[scale=0.5]{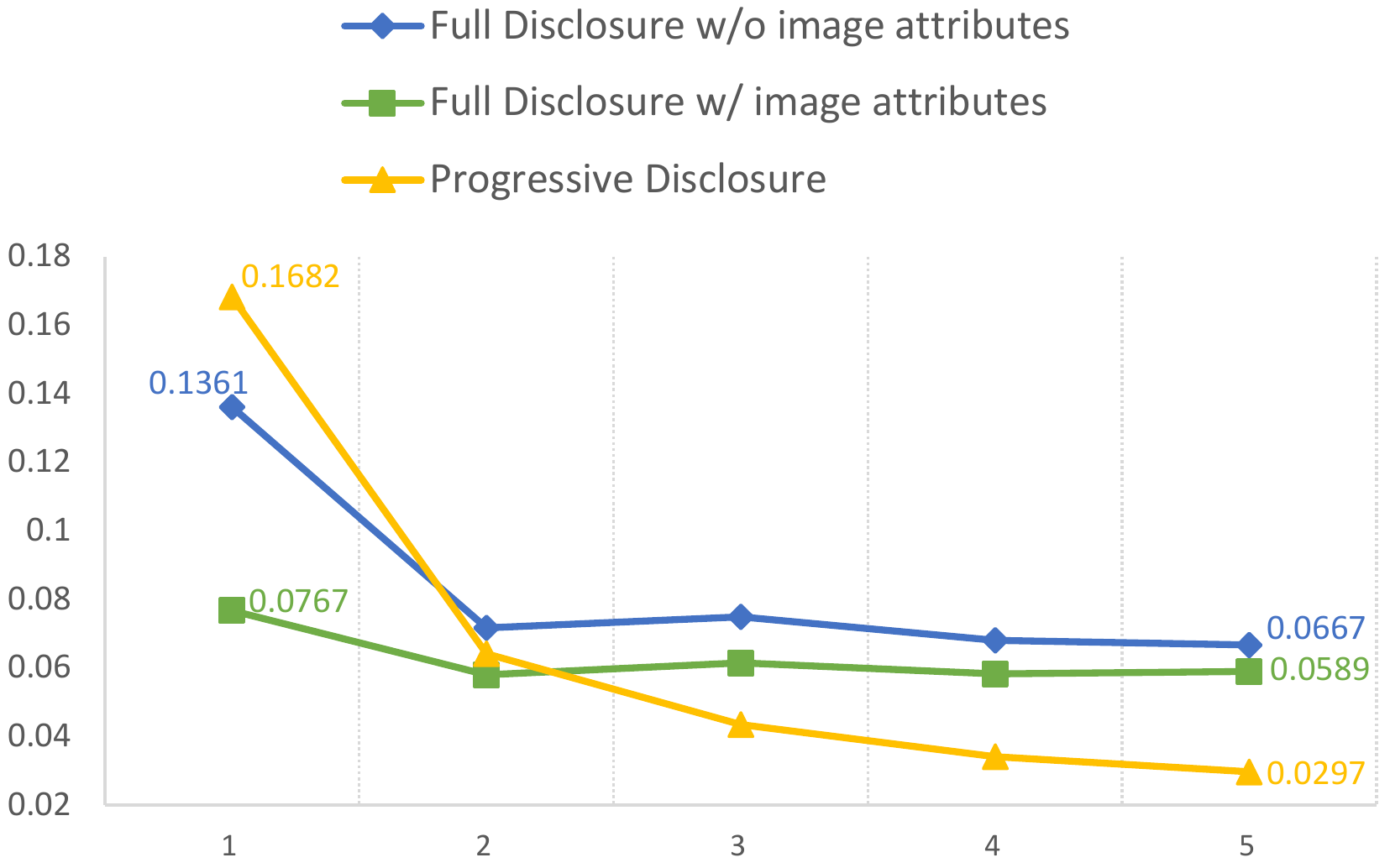}
    \caption{Loss on testing set in 5 rounds under different settings: full disclosure with and without attributes of candidate image and progressive disclosure.}
    \label{exp:LLL}
\end{figure}

While the ranking percentiles are very close between full disclosure and progressive disclosure with image attributes, we can further investigate their performances by referring to their loss curves in Figure ~\ref{exp:LLL}. It is obvious that in the beginning, using progressive disclosure incurs more loss than the other two which matches with their performances at round 1. However, at round 5, the loss for progressive disclosure drops to $0.0297$ while it is $0.0589$, about $1.98$ times as much as the former, for full disclosure with image attributes. Combining the loss with their performances at round 5, we can conclude that using progressive disclosure is more capable of fitting the data.

\subsubsection{Different Ways of Feature Extractions}
Except the different features extracted from various networks, we keep all other settings the same as our best one. In Table~\ref{exp:featext}, there is an apparent superiority in using SE-ResNet on \texttt{CelebA}. The results reveal that the features extracted might have more similar representations with those of attributes, enabling the model to learn their connections well afterwards. 

\begin{table}[h]
    \centering
    \begin{tabular}{c|c|c}
        Features & Percentile & Loss \\
        \hline
        OpenFace    & 87.02\%  &    0.1225         \\
        SE-ResNet on \texttt{VGGFace2}    & 95.80\% &0.0434  \\
        SE-ResNet on \texttt{CelebA}    & \textbf{98.66\%} &   \textbf{0.0297}
    \end{tabular}
    \caption{Ranking percentile and loss are reported on testing set on the final round of the best epoch.}
    \label{exp:featext}
\end{table}

\subsection{Visualizations} %really wanna cry when i get here...fml
\begin{figure*}[!htb]
    \centering
    \includegraphics[scale=0.75]{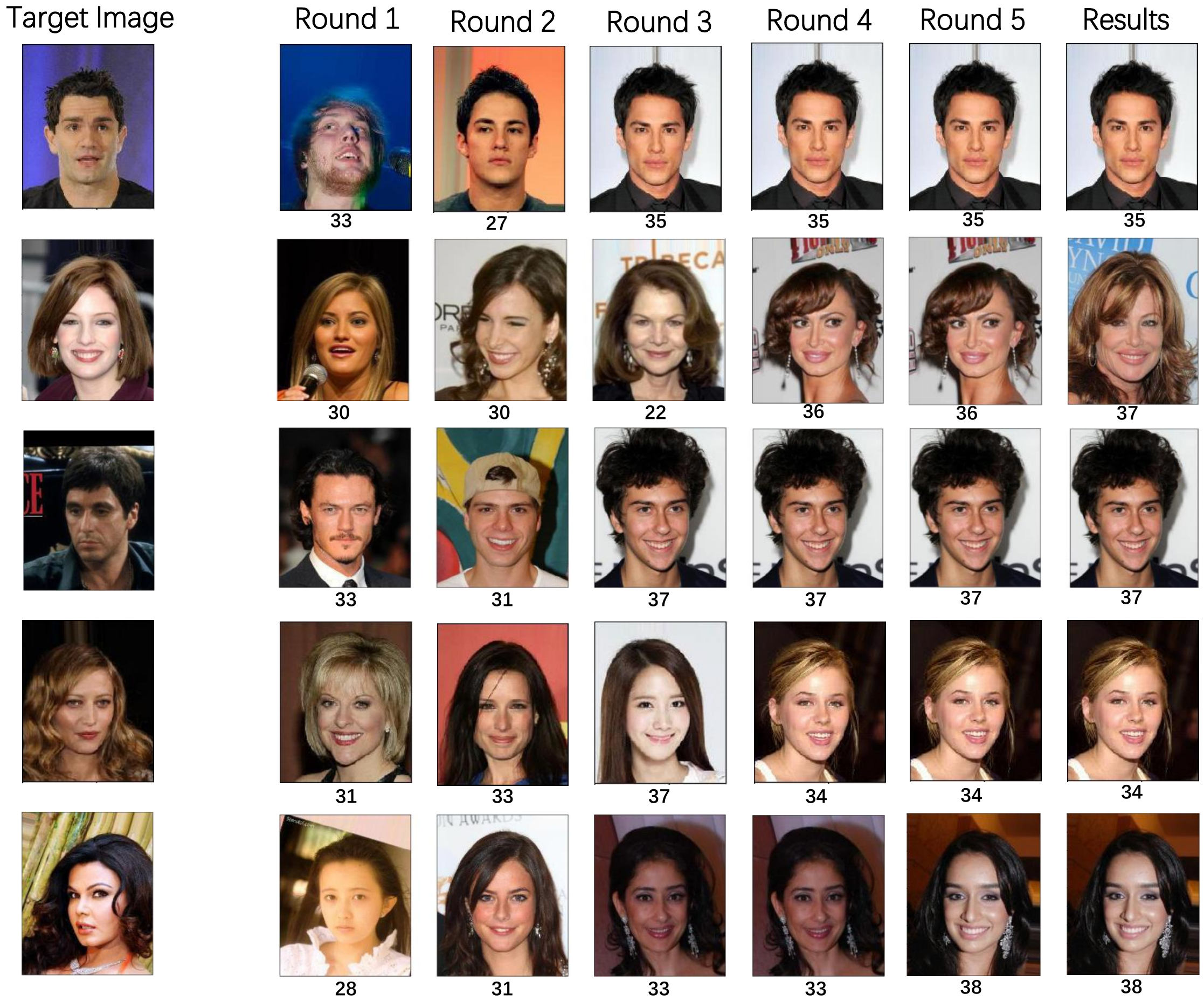}
    \caption{Examples of interactions between users and the proposed facial image retrieval framework with progressive relevance feedback on \texttt{CelebA}.}
    \label{fig:vis}
\end{figure*}
Apart from exhibiting the target image and candidate images during interaction between the system and users, we also calculate the number of attributes that are the same between the target image and the candidate image at each round to provide further investigation from another perspective.

From the visualization results, there are some interesting discoveries. We first find out that for target images of male, the system is more likely to converge at an early stage like at round three (see the first and the third row). However, when it comes to female target images, the system could still change at the last minute after all rounds for a better result (see the second row). These differences might come from the distribution of the dataset, where there are more diversity in female images than male images and it enables the system to refine the retrieval results with finer granularity.

Another interesting thing is that, we might assume that the increase in the number of matched attributes is consistent with the improvements in performance. However, although the number of matched attributes is indeed increasing in most cases, it is not always true. For example, from round 3 to round 4 at row four, the number of attributes drops from $37$ to $34$. Despite that the image from round 4 has $3$ less matched attributes, it definitely seems to be a better match with the target image. At least they all share a western look with blond hair and high cheekbones while the image from round 3 is a typical East Asian face. Likewise, for images from round 2 to round 3 at row two and for images from round 1 to round 2 at row one and row three, the decline in the number of matches attributes essentially leads to a visible better results. This may indicate that our model has the advantage of combining the image features together with attributes information for better retrieval performances.

%\subsection{Perceptual Studies} %real human results
\section{Conclusion}
\label{sec:conclusion}
%\paragraph{Limitations.} 
%\paragraph{Future Work.}
In our work, we shed light on facial image retrieval problem and propose an end-to-end interactive framework with progressive disclosure. We also explore different settings in various scenarios and applications. Though not perfect, our work is sufficient to deal with many cases with over $98\%$ ranking percentile. In the future, this retrieval problem can be upgraded to a conditional generation task that helps suspect sketch. Moreover, the forms of feedback can be expanded and enriched to capture the subtlety, such as fine-grained attributes or even verbal descriptions which enables smoother and more nature approach. The abundant ambiguity in the perception of facial images makes it particularly difficult but crucial to have an intelligent and accurate method to bridging the semantic gap and we hope our work can be a stepping stone for interested researchers.
\section{Acknowledgement}
I am grateful for all the support from Yuwen Xiong on this work. Though not an official author, his encouragements and guidance to me on this work is one of the most important reasons of me finishing this paper.
\cleardoublepage
\newpage
\newpage
{\small
  \bibliographystyle{ACM-Reference-Format}
  \bibliography{main}
}
\end{document}